%% file: main.tex
\documentclass{llncs}

\usepackage{graphicx} 
\usepackage{pdflscape} 
\usepackage{booktabs} 
\usepackage[left=2.5cm,right=3cm,top=3cm,bottom=2.5cm]{geometry}
\usepackage{fancyhdr} 
\usepackage[utf8]{inputenc}
\usepackage{float}
\usepackage{datetime}
\usepackage{natbib}
\usepackage{setspace}
\usepackage{amsmath}
\usepackage{hyperref}
\usepackage{verbatim}
\usepackage{algorithm}
\usepackage{algpseudocode}
\usepackage{marginnote}
\usepackage{rotating}

\usepackage{amsfonts}

\algnewcommand\algorithmicinput{\textbf{Input:}}
\algnewcommand\Input{\item[\algorithmicinput]}
\algnewcommand\algorithmicoutput{\textbf{Output:}}
\algnewcommand\Output{\item[\algorithmicoutput]}
\usepackage{comment}

\usepackage{bbm}
\usepackage{pifont}
\usepackage{subfig}
\usepackage{xr}

\usepackage{xcolor}

\hypersetup{
    colorlinks,
    linkcolor={red!50!black},
    citecolor={red!50!black},
    urlcolor={blue!80!black}
}

\begin{document}

\title{Using Neural Generative Models to Release Synthetic Twitter Corpora with Reduced Stylometric Identifiability of Users}
\titlerunning{Neural Models for Synthetic Text}  
%
\author{Alexander G. Ororbia II \and Fridolin Linder \and Joshua Snoke}
\authorrunning{Ororbia et al.} 
%
%
\institute{Pennsylvania State University, Old Main, State College, PA 16801, USA\\
\email{\{ago109,fridolin.linder,snoke\}@psu.edu}}

\maketitle            

\begin{abstract}
We present a method for generating synthetic versions of Twitter data using neural generative models. The goal is protecting individuals in the source data from stylometric re-identification attacks while still releasing data that carries research value. Specifically, we generate tweet corpora that maintain user-level word distributions by augmenting the neural language models with user-specific components. We compare our approach to two standard text data protection methods: redaction and iterative translation. We evaluate the three methods on measures of risk and utility. We define risk following the stylometric models of re-identification, and we define utility based on two general word distribution measures and two common text analysis research tasks. We find that neural models are able to significantly lower risk over previous methods with little cost to utility. We also demonstrate that the neural models allow data providers to actively control the risk-utility trade-off through model tuning parameters. This work presents promising results for a new tool addressing the problem of privacy for free text and sharing social media data in a way that respects privacy and is ethically responsible.
\end{abstract}

\section{Introduction}
This paper presents a method for producing synthetic tweet corpora that are released in place of the original tweets, with the dual goals of minimizing the identifiability of users and maintaining user-specific word distributions and the analytical usefulness of the released corpus. In the field of data privacy, these two goals, colloquially known as minimizing risk and maximizing utility, see \cite{duncan2001disclosure}, form the basis for measuring the effectiveness of protection methods. Risk and utility reside in tension, such that protection methods can be tuned to offer different levels of utility for corresponding levels of risk. Ideally, we want protection methods that offer the best trade-off of risk and utility in order to allow data providers to release high-quality data.

When dealing with unstructured text documents, privacy risk for users is commonly measured through stylometric models that identify anonymized authors based on features extracted from the text. These models have been successful in correctly attributing authorship with documents such as the Federalists papers; \cite{mosteller1964inference}, online blogs; \cite{narayanan2012feasibility}, and even executable binaries; \cite{caliskan2015coding}. Recent work has shown similar results with short documents such as tweets, despite the small character limit; \cite{almishari2014stylometric}.

We are concerned with a stylometric attack on a corpus of tweets that were gathered and released for research purposes, which is commonly done in the social sciences. This corpus may include tweets that are no longer publicly available, either from users deleting their own content or from Twitter enforcing their policies, e.g., terms of service violations or length of time. As others have pointed out, i.e., \cite{williams2017towards}, Twitter states in their developer guidelines that users have the right to be forgotten, see \cite{twitterUA}. Thus if tweets are deleted, researchers should not use or release those tweets further. However, by complying with this policy, research can be rendered non-replicable since original data cannot be shared; \cite{tromble2017we}. In addition to this, a number of authors have called for stronger ethical standards when publishing social media data, e.g., \cite{williams2017towards,zimmer2010but,rivers2014ethical}. The privacy concern for users comes from either information disclosed through the sharing of previously deleted tweets or through additional information generated from analyses, e.g., identifying hate speech; \cite{burnap2015cyber}. In that case, if users were identified and linked to their tweets, person-specific information would be made public beyond what users had intended to share.

While most of the stylometry literature has focused on developing attack models, some methods do exist to protect the text from stylometric attacks. These include semi-automatic methods to obscure individual authors; \cite{kacmarcik2006obfuscating} or approaches such as iterative translation; \cite{rao2000can,mack2015best}, which have been shown to offer little protection in some cases, see \cite{caliskan2012translate}. In addition to risk concerns, redaction or translation are blunt instruments that do not take into consideration the original language of the text and may remove features which are vital for researchers. We address this by taking a neural-based approach, inspired by the promising results in training neural architectures as simple generative models for natural language processing, e.g., \cite{bengio2003neural,mikolov2010recurrent,sutskever2014sequence,ororbia2017learning,serban2017piecewise}. In particular, neural architectures greatly reduce the amount of human involvement required in extracting the properties of language and (word) distributional information from the original documents. We do not simply use ``off the shelf'' neural methods, but we develop a novel architecture to condition our synthesis model on each user in the corpus. From a utility standpoint, generating synthetic tweets that capture the language of the original texts should outperform redaction or iterative translation. On the risk side, neural models will not precisely replicate the original tweets (it would be severe overfitting if they did), so risk should be reduced by the noise added in the new text generation process.

The contributions of this paper are as follows: (1) a novel neural architecture to efficiently synthesize text based on individual users' language features, (2) the inclusion of a tuning parameter which allows for varying the risk-utility trade-off, (3) defining a privacy risk scenario when sharing Twitter data according to stylometric methods, and (4) an empirical evaluation of different private protection methods with respect to this scenario and the utility of the released data. Our experimental results show that data generated by neural models offer an improved risk-utility trade-off as compared with redaction or iterative translation. They also show that the risk-utility trade-off can be managed by the scaling of the neural model's output logits (using a parameter commonly referred to as the \emph{temperature}). This provides a straightforward privacy tuning parameter, allowing for the release of synthetic tweet corpora with different amounts of utility and risk.

The rest of this paper is organized as follows. Section 2 formally defines our risk and utility measures. Section 3 details our neural synthesis method and prior methods of redaction and iterative translation. Section 4 provides a thorough experiment evaluating the risk and utility offered by different methods on a real Twitter dataset. Section 5 gives a discussion of our results and conclusions.

\section{Problem Formulation}
We wish to release a protected version of an original tweet corpus, and we evaluate different methods of doing so based on empirical measures of risk and utility. We define the risk of such a release as the identifiability of the individuals in the data. We define the utility by the preservation of research value as compared with the original data. In the following sections, we precisely detail how we measure each of these values. These metrics are used for our experimental evaluation in Section \ref{sec:experiment}.

\subsection{Re-identification attack model}
\label{sec:idenmodel}
Assuming the released tweet corpus is ``anonymized", i.e., the real usernames are removed and replaced with an anonymous identifier to connect all of a user's tweets in the corpus, identification can still occur through a stylometric attack, which links publicly available texts known to be composed by the users in the corpus. We assume the tweets do not contain metadata, which would need to be protected by additional methods. We emulate a stylometric attack by using an approach akin to that of \cite{narayanan2012feasibility} and \cite{almishari2014stylometric}. We assume a release dataset, $R$, and an attack dataset, $A$, with matching sets of $k$ users, $U = \{u_1,u_2,...,u_k\}$. For each user in $R$ and $A$, there are a set of corresponding tweets: 
\[
t^R_{ij}, \hspace{5mm} i \in 1...n^R_j \hspace{5mm} \& \hspace{5mm} j \in 1...k
\]
\[
t^A_{ij}, \hspace{5mm} i \in 1...n^A_j \hspace{5mm} \& \hspace{5mm} j \in 1...k
\]
The attacker who wishes to identify users in $R$ collects $A$, which has labeled tweets for the users of interest. In a realistic scenario, the attacker likely does not know the exact set of users in the target data (or is not interested in disclosing all of their identities), but by assuming the user sets of $R$ and $A$ are the same, we place an upper bound on the attacker's knowledge of users in the release data. This allows us to estimate the identification percentage across all users in the sample.
\begin{table*}[!t]
\caption{Feature sets used for the stylometric attack models.}
\label{tab:features}
\centering
    \begin{tabular}{llc}
    \hline
      Feature Set & Description & \# Features \\ \hline
      Uni-grams & Count of all lemmatized uni-grams that occur in at least \\
      & two tweets and in no more than 50\% of the tweets & 100,959 \\
      Bi-grams & Count of all lemmatized bi-grams that occur in at least two \\ 
      & tweets and in no more than 50\% of the tweets & 349,853 \\
      Almishari et. al (2014) & Char-Uni-grams & 26 \\ 
      & Char-Bi-grams & 676 \\
      Narayanan et. al (2012) & Tweet word length, Tweet character length, Vocabulary richness, & \\
      & Word shape, Word length, Char-Uni-grams, Number of & \\
      & punctuation characters, Number of digits, Number of & \\ 
      & special characters (non-ascii characters), Number of \\ 
      & function words, Syntactic category pairs & 387 \\ \hline
\end{tabular}
\end{table*}

For each user in $A$, we train a classifier on $A$ to predict each user versus all other users, resulting in $k$ trained classifiers. We utilize a variety of feature sets for this classification task, so we are testing the risk from multiple different attacks. 
Table \ref{tab:features} displays the five different feature sets used for the attack models. We use the uni-grams and bi-grams as simple common sense feature sets, and we use the feature sets from \cite{almishari2014stylometric} and \cite{narayanan2012feasibility} because they were both shown to perform quite well for re-identification.

For each of these feature sets, we try four classifiers: regularized least squares (RLSC), support vector machines (SVM), naive Bayes (NB), and k-nearest neighbors (KNN). These models were used in either \cite{narayanan2012feasibility} or \cite{almishari2014stylometric}. For RLSC, SVM, and KNN, we use row and feature normalization following \cite{narayanan2012feasibility}. As with the feature sets, it is highly likely an attacker would try multiple different classifiers to achieve re-identification, so we test multiple models in order to calculate an estimate of the risk under different attacks. 

For each classifier and feature set trained on $A$, we test using the data in $R$, producing a probabilistic ranking of user matches from highest to lowest. Our overall identification risk can be summarized as the percentage of users with correct matches in the top $x$ (e.g. 1, 5, 10) most likely users. For example, if $x$ is 1, this is the percentage of most likely users based on our classifier which match the true identity. We also fix the number of test samples for each user ($n^R_j$) in $R$ at 99 tweets in order to have comparable testing sets for each user. As suggested by \cite{narayanan2012feasibility} we collapse the feature vectors for each user in $R$ to the mean for testing. This allows us to gain a single ranking of user classifications rather than a set for each tweet. We do not collapse feature vectors during training, except in the case of the KNN, where we collapse vectors to the centroid for each user to improve computational ability.

\subsection{Evaluating research utility}
\label{sec:utilitysetup}
We measure the utility or usefulness of the released tweet corpus in two ways: (1) the similarity of word distributions and (2) similarity of results from commonly used text analyses. These values are compared to the baseline, i.e., the original, unaltered data, since the maximal utility would come from a researcher having access to the original tweets.

For distributional similarity, we use the cosine similarity of the uni-grams and bi-grams between the original and each altered corpus, defined as:
\[
CosSim(x,y) = \frac{\Sigma_i x_iy_i}{\sqrt{\Sigma_i x_i^2}\sqrt{\Sigma_i y_i^2}}
\]
where $x$ and $y$ are term frequency vectors, either of uni-grams or bi-grams, for a given user in the original and the altered corpus. Because the data has user labels, we calculate the similarity for each user and take the average across all users as our utility measure. We do this because each user in the corpus may have specific interests or ways of communicating, which a researcher would want to utilize. Our protection method should preserve not only the overall word distribution, but it should preserve each user's distribution as well. The cosine similarity takes values between 0 and 1, and an average user cosine similarity closer to 1 implies the protection method has better preserved the overall language of each user.

For model-specific assessment, we consider the results from two common types of analyses. The first is classification task to predict tweets that reference the band ``One Direction'', either by name (`one direction'), hashtag (`\#onedirection') or handle (`@onedirection'). This target was chosen due to the relatively high incidence rate ($\sim20\%$) compared to other terms. We label tweets with these references and train a model on the original corpus to predict the label given the rest of the terms in the tweet. Using cross-validation we get a baseline estimate of the out-of-sample F1-score, which combines precision and recall:
\[
\text{F1-score} = \frac{precision * recall}{precision + recall}.
\]
We then test the cross-validated model on each protected dataset to get F1-scores for each of the altered corpora. We expect the perturbation, either from redaction, translation, or synthesis to decrease the accuracy of the model. Generally speaking, the better the release data maintains the feature distribution of the baseline data, the closer the resulting F1-score should be to the baseline F1-score. Another way to think about this is that if the altered corpus carried the same distributional properties as the original corpus, we would expect similar performance on a test set to that of cross-validation.

For the second model-specific measure, we analyze how well the different protection methods preserve sentiment by user, using the ``Vader'' sentiment model, specifically developed to measure sentiment in social media data; \cite{hutto2014vader}. For each release corpus, we generate a vector of user sentiment scores, representing the average compound sentiment (across tweets) for each user. The utility measure is then the cosine similarity between the vector for each release and the baseline's vector. While this is similar in nature to the general word distribution measures, building sentiment models is a very common analysis task using Twitter data. If we were working with numerical data this would be similar to comparing the covariance matrices versus comparing a specific regression model. While the former tells us overall how the datasets compare, the second gives us a specific insight into how the protection methods affect our inference models. Before moving to the empirical evaluation, we detail our proposed protection method in the next section.

\section{Protection methods} 
\subsection{Synthetic Data Models for Privacy Protection}
\label{sec:synData}
The concept of synthetic data originated out of the multiple imputation literature, see \cite{rubin1993}, based on the simple concept of drawing new “samples" from a Bayesian posterior predictive distribution (BPPD) to release in place of the original data. If modeled well, synthetic data capture distributional aspects of the original data resulting in high utility, and none of the values are the original ones ensuring low risk. For general background and further readings see \cite{RRR_2003,reiter_partial,reiter2005using,reiter_kinney_2012,Drechsler_book,raab2017practical}.

Traditional synthetic data is unfortunately not realistic in the scenario we present in this paper. These generative models are either approximated jointly or using fully conditional sequential models. Both in terms of producing accurate synthetic data and running the computation in reasonable time, neither of these present a viable option for unstructured text data. For example, a standard synthesis method for our dataset would require estimating a Multinomial distribution with over 40,000 parameters at a minimum. 

As current synthesis methods do not exist to handle this type of data, we propose a neural-based approach to address this problem. Specifically, to avoid paying a high cost in fitting models that scale poorly to high-dimensional data, we take advantage of a specific class of neural architectures known as recurrent neural networks (RNNs), which are powerful function approximators, see \cite{goodfellow2014generative,chung2015recurrent,sonderby2016ladder}, that process data sequentially and share parameters efficiently through a recursive formulation. This is ideal for text because we can efficiently approximate models that take sequence into account, which is crucial for modeling grammar and syntactic structure in language. In addition, we can take advantage of recent advances in GP-GPU hardware used in speeding up the many matrix operations that underly inference and training in neural networks. 

\subsection{A Multi-User Conditional Synthesis Model}
\label{sec:model}
We aim to model the unknown data generating process underlying an observed tweet corpus, $R$, conditioned on the set of users, $U$. To do this conditioning, we propose a structural modification to the standard recurrent neural architecture. We allocate a set of parameters to model user-specific information and a set of parameters that contain input symbol information, and, to better capture information over long sequences, we integrate a longer-term memory mechanism into the model's hidden state function by creating a model based on the Differential State Framework proposed in \cite{ororbia2017learning}. This model we will call the multi-user Delta-RNN. Along with this model, we explore in our experiments this same structural modification applied to more standard RNN models that also fall under this framework, yielding the multi-user Gated Recurrent Unit (GRU) and the multi-user Elman RNN.

In order to build a generative model, we need to break our unstructured text into a set of features. Rather than utilize a common Bag-of-Words (BOW) representation of our data, which leads to several hundreds of thousands of features, we use a character level feature set, which results in a dramatically smaller dimensionality. We can produce sensible text using character features, since these models take sequence into account. The character set we use includes the 26 standard characters of the alphabet and the integers $0-9$ as well as punctuation and other non-standard symbols, such as emoticons (or emojis).
Synthesizing a tweet involves sampling the learned model's output to generate a sequence of characters until a certain stopping criterion is reached. In our case a sequence is continuously generated until either a simple end-token, e.g., $[eos]$, is generated
or an upper bound on the character limit is reached (this is particularly useful for Twitter data, which caps the total characters at 140 in our corpus).

One key feature of the architectures we develop is that they capture the dependence between documents generated for the same user. Formally, given a finite dictionary of characters $D$, we assume that for input sequences $\mathbf{c} = (\mathbf{c}_{t},\mathbf{c}_{t-1},\dots,\mathbf{c}_{t-m})$ the distribution of interest is $p(\mathbf{c}_t|j,\mathbf{c}_{<t})$, where $\mathbf{c}_t$ is a 1-of-$|D|$ encoding of a character at time $t$ indexed in $D$,\footnote{Note that, in line with the literature, we refer to the position of a character in a sequence or document as temporal information.} $\mathbf{c}_{<t}$ is the historical context---that is, earlier characters in the sequence---preceding the current character of length $m-1$ and $j$ is the index of the user to be associated with this encoded character. By learning a good generative model for this conditional distribution, we can generate valid character sequence samples representative of specific target users. In the architecture proposed in this paper, this can be easily done by feeding a 1-of-$k$ encoding of the user index.

\begin{figure*}[!t]
\centering
\includegraphics[width=0.8\textwidth]{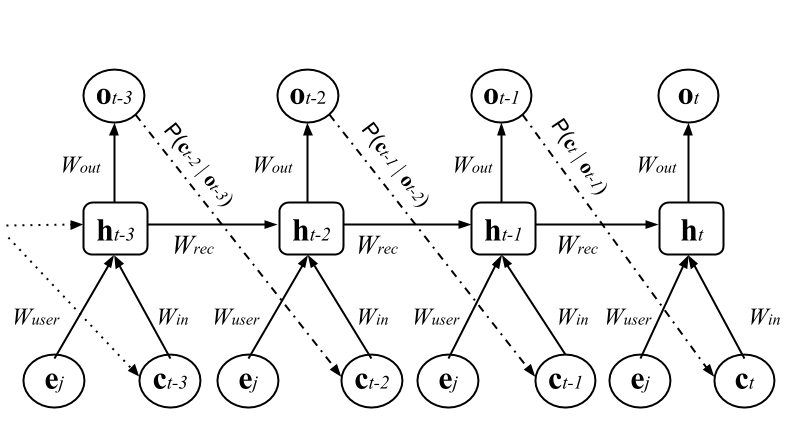}
\caption{The proposed user-based neural generative architecture, unfolded over four time steps. Note that the model depicted here features a simple recurrence relation, but can be adapted to use gating. The user-id vector is repeated over the length of the tweet character sequence. $\mathbf{e}_j$ is the 1-of-\emph{k} encoding of the user index $j$.}
\label{fig:neural_arch}
\end{figure*}

In order to ultimately generate sequences of characters, our recurrent neural architectures specifically learn to approximate $p(\mathbf{c}_t|j,\mathbf{c}_{<t})$. This learning procedure entails processing two parallel streams of encodings--a stream of characters and a corresponding stream of user indices (injectively mapped). The model predicts the characters at time $t+1$ conditional on past characters in the sequence and the user for which the document is generated. This is achieved by using a vector summary of the past (i.e., the last hidden state $\mathbf{h}_{t-1}$), the currently observed token at $t$, and knowledge of the current user $\mathbf{e}_j$ (also a 1-of-k binary encoding corresponding to the user $u_j$). The task is to fit this model to the target corpus $R$ that we want to synthesize.

In contrast to building a separate model for each user in the data-set, our model shares its ``language model'' parameters across the multiple views of the data, motivated by the design of the hybrid architectures of \cite{ororbia_deep_hybrid_2015a,ororbia2015learning,ororbia2015online} or the log-linear models of \cite{le_distributed_2014}. In this way, we may construct a single model that efficiently learns user-specific parameters (contained within a single $|H| x |R|$ matrix) jointly with the general language model parameters (which aggregates across users). This process emulates the sharing of parameters in more classical hierarchical models.

Interestingly enough, our specialized neural architecture also addresses a recent problem found in neural conversation agents; \cite{vinyals_neural_2015}. This issue centers around ``coherence'', where a trained neural model has no sense of identity or self (even at the crudest level). For example, if asked the question, ``Are you married?'', the model responds with ``No!'', but when followed up with the question ``What is your wife's name?'', the model might respond with, ``Cynthia''. While this issue is more prominent in sequence-to-sequence modeling tasks (as in question-answering, dialogue modeling), we also argue that in synthetic data-generation, where samples often come with meta-data, having a model that preserves local information, such as user identity, is crucial. We note that other work has attempted to address the coherence/consistency problem in neural models; \cite{sordoni_neural_2015,li_persona_based_2016}. However, our approach is notably simpler in representing user-local information and differs in problem context compared to sequence-to-sequence mapping models. 

Formally, we write the model parameters $\Theta = (W_{user},W_{in},W_{rec},W_{out})$ (biases have been omitted for clarity). For the multi-user Elman RNN, we calculate hidden and output states via the following equations:
\begin{align}
\mathbf{h}_t &= \phi_{hid}(W_{in} \mathbf{c} + W_{user} \mathbf{e}_j + W_{rec} \mathbf{h}_{t-1}) \label{h_given_x}\\
\mathbf{o}_t &= \phi_{out}(W_{out} \mathbf{h}_t) \label{o_given_h}
\end{align}
where $\mathbf{e}_j$ is the one-hot encoding of a user index $j$ and $(\phi_{hid},\phi_{out})$ are the activation functions for input-to-hidden and hidden-to-output layers respectively. The architecture specified by the above equations, unfolded over time, is shown in Figure \ref{fig:neural_arch}. 
A more powerful variation of the multi-user structure presented above is what we call the multi-user Delta-RNN. This model is defined by the following set of equations:
\begin{align}
\mathbf{d}^{rec}_t &= V_d \mathbf{h}_{t-1}, \quad \mathbf{d}^{dat}_t = W \mathbf{x}_t + W_{user} \mathbf{e}_j,\\
\mathbf{d}^1_t &= \mathbf{d}^{rec}_t \otimes \mathbf{d}^{dat}_t,\ \mathbf{d}^2_t = \mathbf{d}^{rec}_t + \mathbf{d}^{dat}_t,\\
\mathbf{z}_{t} &= \phi_{hid}(\mathbf{d}^1_t + \mathbf{d}^2_t), \\
\mathbf{h}_{t} &=  \Phi( (1 - \mathbf{r}) \otimes \mathbf{z}_{t} + \mathbf{r} \otimes \mathbf{h}_{t-1} ) \mbox{,  and,} \\ \mathbf{r} &=  1 / (1 + exp(-[\mathbf{d}^{dat}_t + \mathbf{b}_r]))
\end{align}
where $\otimes$ means elementwise multiplication and $\mathbf{r}$ is the interpolation gate that the model uses to control how much new information from text at any time step is mixed into the slowly-changing hidden memory state; \cite{ororbia2017learning}.
In Appendix \ref{appendix_a}, we give specifications for regularizing these models through layer normalization \cite{ba2016layer}, as well as a description of the third model, the multi-user GRU. 

For $\phi_{out}$ we choose to parametrize the posterior as a maximum-entropy classifier, defined as follows: 
\begin{align}
P(w,\mathbf{h}_t) = P_{\Theta}(w|\mathbf{h}_t) =  \frac{\exp{( w^{\text{T}} W_{out} \mathbf{h}_t / \tau)}}{\sum_{w'} \exp{((w')^{\text{T}} W_{out} \mathbf{h}_t / \tau)}}, \label{max_ent}
\end{align}
where $\tau$, the temperature meta-parameter, controls the smoothness of the output probability distribution. When $\tau$ is increased (i.e., $\tau \rightarrow \infty$), the output classes tend towards equal probabilities, yielding more uniform samples 
, increasing the risk of mistakes but lowering the probability of generating text snippets similar to the original text. Lowering $\tau$ (i.e., $\tau \rightarrow 0^{+}$) sharpens the posterior distribution, making the probability of the target class, the class to which the model attributes highest probability, closer to 1. This means that the model will be more driven by the frequently observed terms from the original data for each user,
but its samples will be made with higher confidence. At too low of temperatures we may only obtain high-frequency samples, chopping off the tail of the observed distribution of 
characters. We expect that raising the temperature will lower the distributional similarity of users to the original data (harming utility), but will lower the identification risk. 


Parameters of the model are fit to the data via stochastic gradient descent using truncated back-propagation through time to calculate parameter gradients, where $m$ is used to control the length of the window, or number of steps back in time. The objective is to minimize the negative log likelihood of the predictive posterior over the sequence as follows:
\vspace{-0.15cm}
\begin{equation}
\label{objective}
\mathcal{L}(\mathbf{c}) = -\sum^{m}_{t=1} \log p(\mathbf{c}_t|\mathbf{o}_{t-1}) \mbox{.}
\end{equation}
\vspace{-0.45cm}

To generate samples from the neural model, we simply make use of the model's efficient inference procedure, similar to that in \cite{graves_generating_2013}. Specifically, by clamping the input units corresponding to a desired target user index and feeding in a  ``null'' vector (or vector of all zeroes) as initial input,
 we may sample from its output probabilities and ultimately generate synthetic symbol sequences for individuals by feeding in a sample of model's predicted output back in as input for the next step. 

\subsection{Previous protection methods}
\label{sec:prevProtection}
Two other protections method were applied to the release data 
set, redaction, and iterative translation. For redaction, we simply removed all hashtags and handles present in the corpus, assuming that many users frequently repeat hashtags and handles which makes these features best used for identifying a user. The iterative translation was based on \cite{mack2015best} and worked by first translating the original (redacted) corpus into Arabic and then back into English using the Google Translate API\footnote{\url{cloud.google.com/translate}}. Arabic was chosen because it was found to offer the highest level of protection in \cite{mack2015best}. The translation was performed on top of the redacted data because the hashtags and handles would likely not change through the translation process.

\section{Experiment}
\label{sec:experiment}
\subsection{Dataset}
\label{sec:data}
Our experimental dataset consists of tweets collected by \cite{barbera2016governments} for a study on German Twitter users. Twitter users were randomly sampled by generating uniformly random Twitter IDs and selecting based on profile-level selection criteria. We subset the data selecting only tweets classified as English by Twitter's language detection algorithm. We then divided the corpus into two parts, as described in Section \ref{sec:idenmodel}. The attack dataset, $A$, contained 386,684 tweets across 627 users and the release dataset, $R$, contained 62,073 tweets across the same 627 users.

\subsection{Experimental Design}
\label{sec:experimentaldesign}  
We produced 15 total neural generated synthetic datasets using three model variations and five temperature settings as described in Section \ref{sec:model}. The three variations of our neural synthesis model were the Elman RNN, the Delta-RNN, and the GRU. We optimized parameters through stochastic gradient descent using the ADAM \cite{kingma2014adam} adaptive learning rate scheme, and a step size of $\lambda = 0.002$ was found to be sufficient in preliminary experiments. All models were trained for 150 epochs, with updates calculated using mini-batches of 32 samples. Since these tweets are bounded by the 140 character limit, we were able to use full back-propagation through time without any truncation. Regularization consisted of layer normalization; \cite{ba2016layer}. See Appendix \ref{appendix_a} for technical details of these model architectures.
For each of the three models, we produced synthetic corpora for each of the temperatures: $\{0.25, 0.5, 1, 1.25, 1.75\}$.

The re-identification risk was estimated following the stylometric attack models given in Section \ref{sec:idenmodel}. There are five feature sets and four classification models, resulting in a total of 20 risk estimates. The utility was estimated using the four measures detailed in Section \ref{sec:utilitysetup}. The two general measures were average user uni-gram and bi-gram cosine similarity between the baseline data and each altered dataset, with a high similarity implying most users have similar word distributions between the baseline and altered datasets. The two model-specific measures were a classification task to predict a reference to the band ``One Direction'' and a sentiment analysis task. High utility implied similarity of classification or sentiment between the original and synthesized corpora.

\subsection{Experimental Results}
\label{sec:expResults}
\begin{figure*}[!t]
\centering
\includegraphics[width=0.9\textwidth]{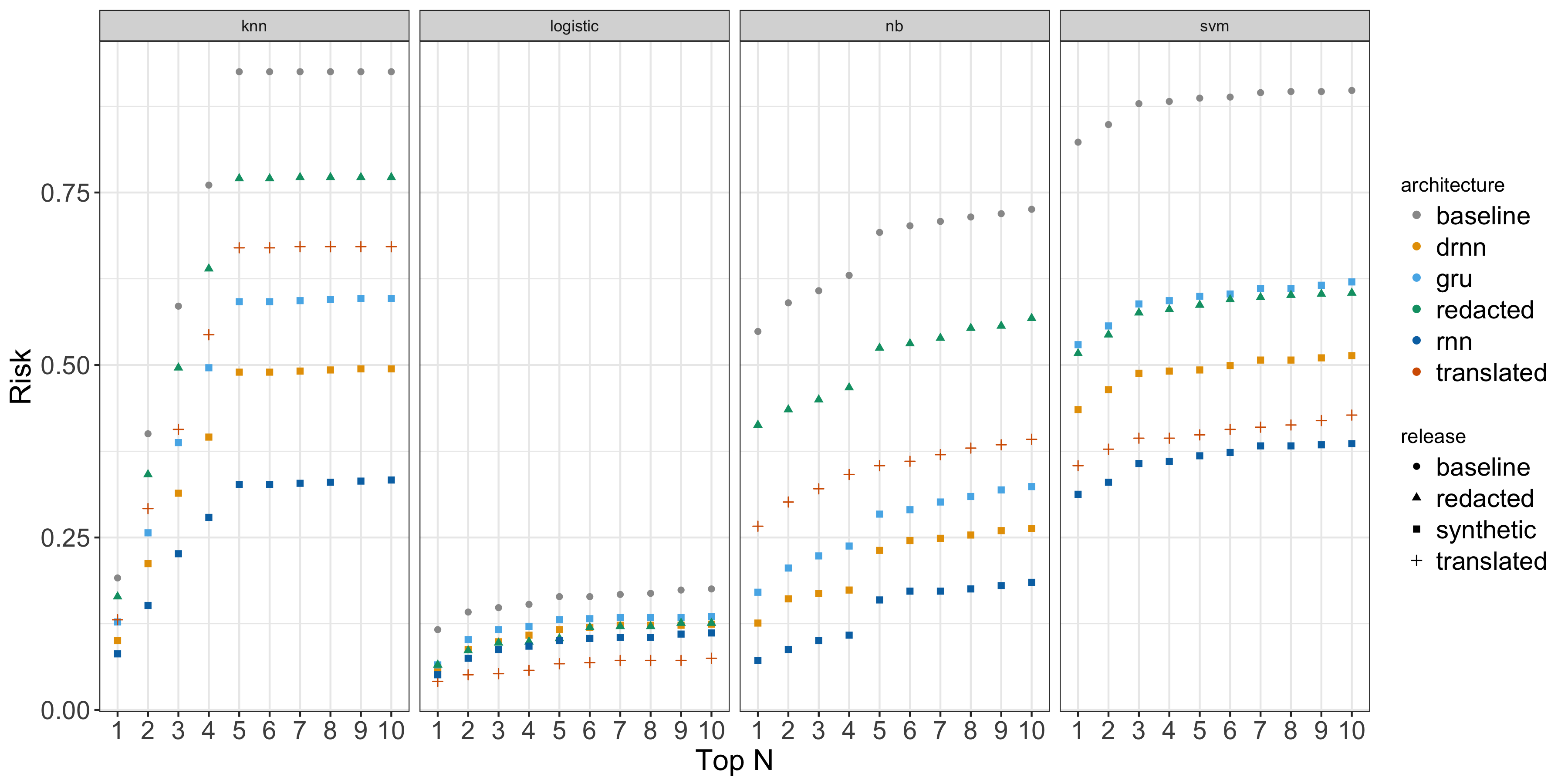}
\caption{Re-identification risk using the bi-gram feature set for four models (KNN, RLSC, NB, SVM). Y-axis denote the percentage of users correctly identified, and the x-axis denotes whether it is for the top 1, 2, ..., 10 most likely users.}
\label{fig:risk_bigram}
\end{figure*}

Considering the stylometric attack results on the baseline data, we find that the bi-gram features show the best performance across the five feature sets, and for that feature set the SVM model performed the best (highest percentage of users identified as top 1 most likely in the baseline). Using the bi-gram SVM, the true user was correctly identified as the most likely candidate for 82\% of users. This shows a very significant risk of re-identification in the unaltered data. These re-identification levels are comparable to previous work; \cite{narayanan2012feasibility,almishari2014stylometric}.

\begin{figure*}[!t]
\centering
\includegraphics[width=\textwidth]{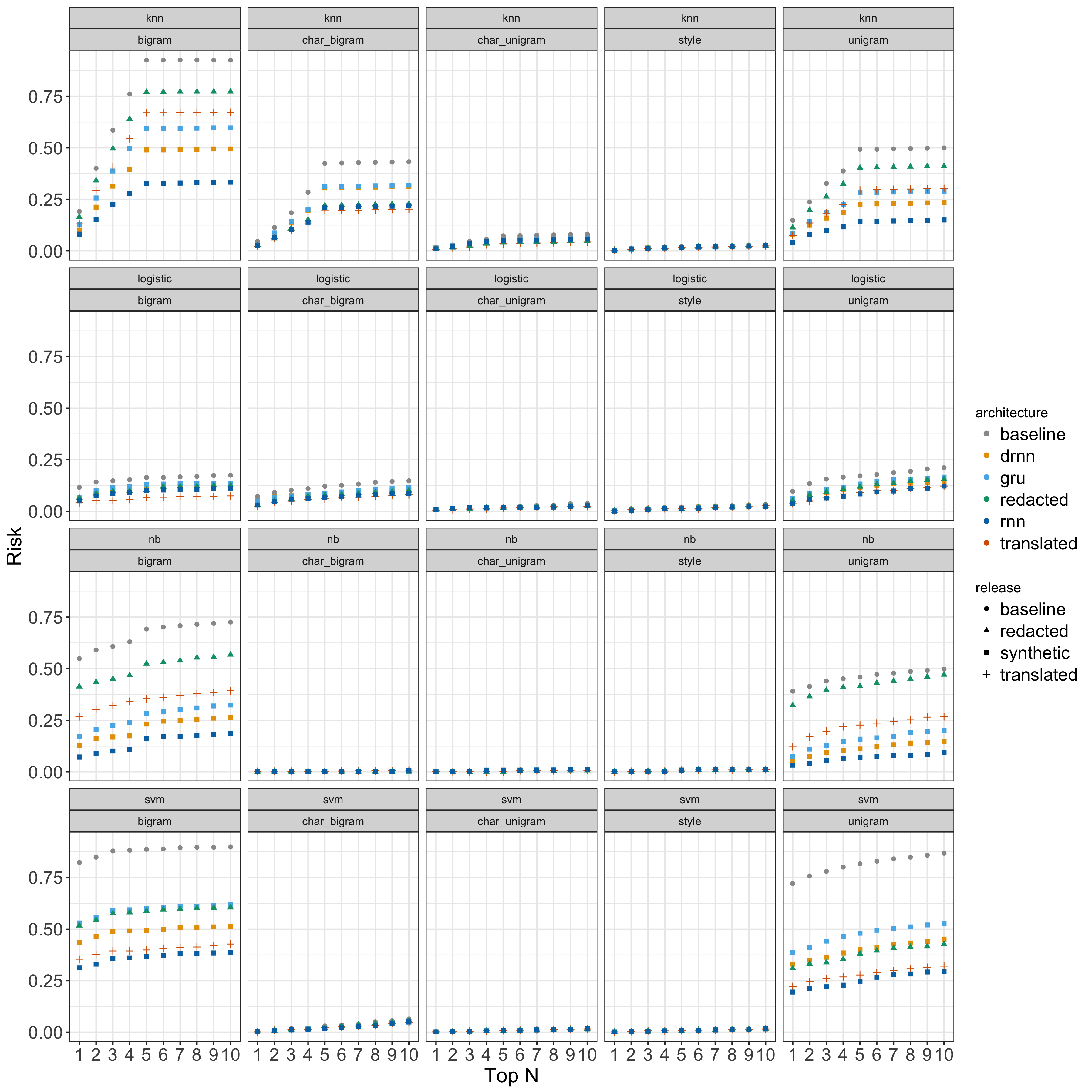}
\caption{Re-identification risk using the five feature sets (uni-gram,bi-gram, character uni-gram, character bi-gram, and stylometric) for four models (KNN, RLSC, NB, SVM). Y-axis denote the percentage of users correctly identified, and the x-axis denotes whether it is for the top 1, 2, ..., 10 most likely users.}
\label{fig:risk_all_features}
\end{figure*}

Figure \ref{fig:risk_bigram} shows the risk results for the baseline, redacted, translated, and the three neural models at temperature 1, for each of the four attack models using the bi-gram feature set. The plots show the percentage of users correctly identified in the top $N$ predicted users. We see that for three of the four attack models, the classifiers do quite well at identifying the true user among the most likely. All of the protection methods decrease the risk, but they do so at varying levels. The Elman RNN models has the lowest risk, while either the redacted or GRU have the highest risk out of the set of protection methods. A full plot of the 20 risk results (5 feature sets x 4 models) is shown in Figure \ref{fig:risk_all_features}.

\begin{figure*}[!t]
\centering
\includegraphics[width=0.9\textwidth]{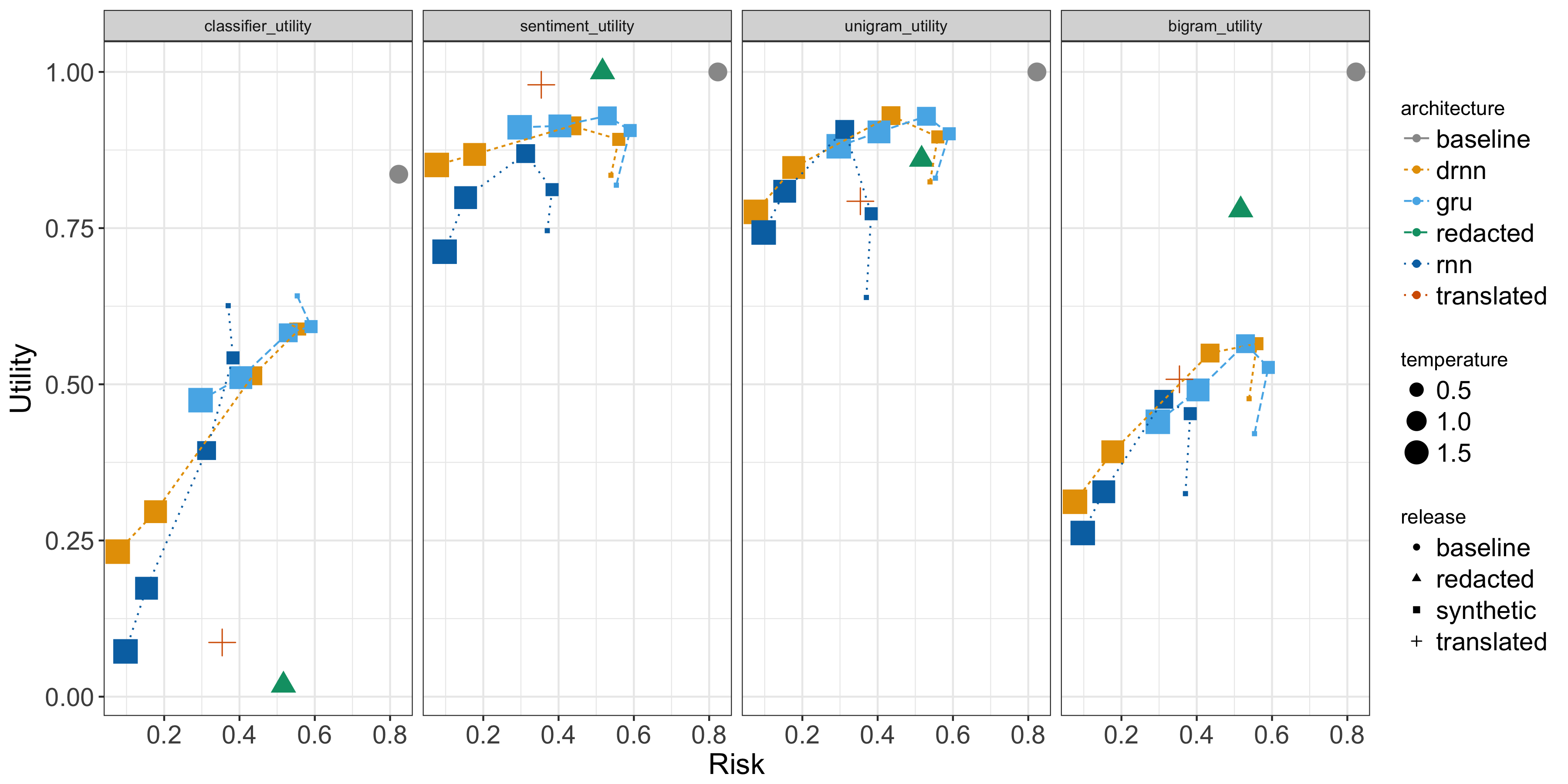}
\caption{Four risk utility maps, one for each utility measure.}
\label{fig:ru_plot}
\end{figure*}

We visualize the risk along with the utility results by using the standard risk-utility mapping, shown in Figure \ref{fig:ru_plot} (full results also given in Table \ref{tab:risk_util_results}). Each plot in the figure shows a risk-utility map for one of our four utility measures. In all cases, utility ranges from zero to one with zero being the lowest utility and one being the highest. Similarly, risk ranges from zero to one with zero being the lowest risk and one being the highest. 
We expect methods to move roughly along a curve from the bottom left corner (no utility, no risk) to the top right corner (all utility, all risk). We use the baseline original corpus release as the standard for maximal utility and maximal risk.
And as expected, the baseline resides generally in the top right corner of the risk-utility maps. Figure \ref{fig:ru_plot} also shows the neural syntheses for each of the five temperatures, plotted on curves as the temperature changes. We see that, apart from a couple exceptions, as the temperature shrinks both the risk and utility increase. This confirms our expectation that the temperature acts as a risk-utility tuning parameter.

The neural syntheses perform quite well for the classification utility task (first panel), while the redacted and translated sit well below the neural curves, implying a better risk-utility trade-off from the neural methods. At very low temperatures the risk still decreases but the utility actually improves, moving the line towards the top left corner. While this is only one example, this is a very promising result. 
Both the redaction and translation methods perform well on the sentiment task (second panel), sitting above the curves formed by the neural syntheses. Overall every method offers very high utility, so the sentiment task gives us less ability to differentiate between the protection methods.

\input{risk_util_results.tex}

For the uni-grams similarity (third panel), the redacted and translation methods fall below the line formed by the neural methods, which implies they offer a worse risk-utility trade-off. For the bi-grams similarity (fourth panel), the translation approach is comparable to the neural methods while the redacted sits above the line. For both general utility measures, the Delta-RNN has the highest curve, offering the best risk-utility trade-off.

For the sentiment, uni-grams, and bi-grams we see that both risk and utility increase as the temperature decreases until the temperature gets very small (0.25). This phenomenon is because at lower temperatures the models start chopping off more from the tails of the user word distributions. This implies both lower risk due to the increased rarity of identifying words and also lower utility due to less distributional similarity. 

\input{example_tweets.tex}

At lower temperatures the more frequently observed terms are more likely to be generated, thus producing tweets that look more like the original language. Table \ref{tab:example_tweets} shows example tweets from the Delta-RNN for each of the temperatures. It is easy to see that the tweets become less like actual English at higher temperatures (though some may wonder if Twitter is ever actual English). Particularly the spelling, since we build character-level models, and the grammar get noticeably worse at higher temperatures. These results show how the temperature parameter can be used as a tool to control the amount of noise introduced into the release data and enables the user to actively control the risk-utility trade-off. This possibility makes the neural protection methods attractive in comparison to the translation and redaction methods that do not allow such flexibility in the choice of trade-off.

\section{Discussion}
\label{sec:discussion}
This paper presents a novel method using neural models for reducing the risk from stylometric attacks on text data while preserving the original language. We show that a more informed risk-utility trade-off is possible over previous methods such as redaction and translation. Particularly we find that the neural methods can reduce the risk far beyond the other methods, at some cost of utility. In some cases, such as uni-gram distributions or the classification task, the neural methods offered a better trade-off, while in the case of bi-grams or the sentiment task they did not. 

When using the temperature (of the output distribution) as a ``knob" to control the risk and utility, the neural models allow us to set the appropriate levels. Redaction on the other hand, while maintaining high utility, allowed over 50\% re-identification of the users which may be unacceptably high. By testing a variety of attack models and feature sets, we showed that our protection methods are robust to more than one attack. In particular, we found a similar level of re-identification risk for the baseline corpus as was shown in previous papers, validating our stylometric attack models.

The use of temperature to control the risk and utility should be further explored in future work. Additionally, we see a different curve for the temperature when using different models. For example, the GRU carried higher risk even for higher temperatures, whereas the Delta-RNN and RNN changed quite a bit as the temperature lowered. In all cases, the desired risk-utility trade-off is important, and giving a range of output possibilities is key. In application, a data provider would need to assess the desired levels of risk and utility and would need tools to match these levels.

In order to further improve the performance of the neural synthesis models proposed in this paper, future work should include reformulating our models such that neural variational inference \cite{kingma2013auto} is employed. This would facilitate the learning of richer probabilistic language models that might capture yet more complex features of the original corpora distributions one might want to synthesize. Furthermore, operating under such a Bayesian framework would allow us to easily integrate better text-specific prior distributions, such as the piecewise-constant distribution \cite{serban2017piecewise}, easing the learning of difficult, multi-modal distributions.

{\footnotesize \bibliographystyle{chicago}
\bibliography{new_ref}}

\input{appendix_A.tex}


\end{document}

%% file: risk_util_results.tex
\begin{table*}[!t]
\caption{Risk and utility for all releases. Risk is the proportion 
           correctly identified as the top 1 classification based on a SVM 
           classifier using bi-gram features.} 
\label{tab:risk_util_results}
\centering
\begin{tabular}{llrrrrr}
  \hline
  Release & & Risk & \multicolumn{4}{c}{Utility} \\
  \hline
  & \vline & Top 1 & Classifier & Uni-gram & Bi-gram & Sentiment \\ 
  \hline
  Baseline & \vline & 0.82 & 0.84 & 1.00 & 1.00 & 1.00 \\ 
  Redacted & \vline & 0.52 & 0.02 & 0.86 & 0.78 & 1.00 \\
  Translated & \vline & 0.35 & 0.09 & 0.79 & 0.51 & 0.98 \\
  Synthetic Delta-RNN (0.25) & \vline & 0.54 & 0.59 & 0.82 & 0.48 & 0.83 \\ 
  Synthetic Delta-RNN (0.5) & \vline & 0.56 & 0.59 & 0.90 & 0.56 & 0.89 \\  
  Synthetic Delta-RNN (1) & \vline & 0.44 & 0.51 & 0.93 & 0.55 & 0.91 \\ 
  Synthetic Delta-RNN (1.5) & \vline & 0.18 & 0.30 & 0.85 & 0.39 & 0.87 \\ 
  Synthetic Delta-RNN (1.75) & \vline & 0.08 & 0.23 & 0.78 & 0.31 & 0.85 \\
  Synthetic GRU (0.25) & \vline & 0.55 & 0.64 & 0.83 & 0.42 & 0.82 \\ 
  Synthetic GRU (0.5) & \vline & 0.59 & 0.59 & 0.90 & 0.53 & 0.91 \\ 
  Synthetic GRU (1.5) & \vline & 0.40 & 0.51 & 0.90 & 0.49 & 0.91 \\ 
  Synthetic GRU (1.75) & \vline & 0.30 & 0.47 & 0.88 & 0.44 & 0.91 \\ 
  Synthetic GRU (1) & \vline & 0.53 & 0.58 & 0.93 & 0.56 & 0.93 \\   
  Synthetic RNN (0.25) & \vline & 0.37 & 0.63 & 0.64 & 0.33 & 0.75 \\ 
  Synthetic RNN (0.5) & \vline & 0.38 & 0.54 & 0.77 & 0.45 & 0.81 \\ 
  Synthetic RNN (1) & \vline & 0.31 & 0.39 & 0.91 & 0.48 & 0.87 \\ 
  Synthetic RNN (1.5) & \vline & 0.15 & 0.17 & 0.81 & 0.33 & 0.80 \\ 
  Synthetic RNN (1.75) & \vline & 0.10 & 0.07 & 0.74 & 0.26 & 0.71 \\ 
  \hline
\end{tabular}
\end{table*}

%% file: example_tweets.tex
\begin{table*}[!t]
\caption{Sample tweets from the Delta-RNN model, showing the progression as the tweets change.} 
\label{tab:example_tweets}
\centering
\begin{tabular}{ll}
  \hline
  Temperature & Example tweet \\
  \hline
  0.25 & \footnotesize\verb|RT @karishmore: Thanks for this more members in my favorite featuring me| \\
  0.5 & \footnotesize\verb|RT @camerondallas: Love the first thing to be with my bestfriend| \\
  & \footnotesize\verb|https://t.co/pwranhzpro| \\
  1 & \footnotesize\verb|These when Same days, but I learned it will tell be same #future #10000.| \\
  1.5 & \footnotesize\verb|Idor’7 series. HappyHappyBirmin| \\
  1.75 & \footnotesize\verb|#Nh0011015. One has micror liden by did masqs i awake imemiquz.| \\
  & \footnotesize\verb|The hunneos\matesr Goz\| \\
  \hline
\end{tabular}
\end{table*}

%% file: appendix_A.tex
\section{Appendix: Model Details}
\label{appendix_a}
In this appendix, we describe in detail, for reproducibility, the variations of the conditional synthesis model trained in this study. Both models have been shown to be special cases that can be derived from the Differential State Framework \cite{ororbia2017learning}.

Layer normalization \cite{ba2016layer}, as applied to a linear pre-activation $\mathbf{z}_t$, is defined simply as follows:
\begin{align}
LN(\mathbf{z}_t) &= \frac{\mathbf{s}}{\sigma_t} \otimes (\mathbf{z}_t - \mu_t) + \mathbf{b}),\\ 
\mu_t &= \frac{1}{H}\sum^{H}_{i=1} a_{i,t},\ \sigma_t = \sqrt[]{\frac{1}{H}\sum^{H}_{i=1} (a_{i,t} - \mu_t)^2}
\end{align}
where $\mathbf{s}$ (the scale) and $\mathbf{b}$ (the shift) are learnable parameters. $\otimes$ denotes the Hadamard product.

\subsection{The Multi-User Delta-RNN}
\label{drnn_model}
While a simple RNN usually requires the layer normalization to be applied after calculating the full linear pre-activation (a sum of the filtration and the projected data point), a Delta-RNN requires further care to ensure the correct components are normalized without damaging the favorable properties inherent to its multiplicative gating.
\begin{align}
\mathbf{d}^{rec}_t &= LN(V_d \mathbf{h}_{t-1}),\\ \mathbf{d}^{dat}_t &= LN(W \mathbf{x}_t) + LN(W_{user} \mathbf{e}_j),\\
\mathbf{d}^1_t &= \mathbf{d}^{rec}_t \otimes \mathbf{d}^{dat}_t,\ \mathbf{d}^2_t = \mathbf{d}^{rec}_t + \mathbf{d}^{dat}_t,\\
\mathbf{z}_{t} &= \phi_{hid}(\mathbf{d}^1_t + \mathbf{d}^2_t), \\
\mathbf{h}_{t} &=  \Phi( (1 - \mathbf{r}) \otimes \mathbf{z}_{t} + \mathbf{r} \otimes \mathbf{h}_{t-1} ) \mbox{,  and,} \\ \mathbf{r} &=  1 / (1 + exp(-[\mathbf{d}^{dat}_t + \mathbf{b}_r])) \mbox{.}
\end{align}
In this paper, we set $\Phi(v) = v$, the identity, and $\phi_{hid}(v) = tanh(v)$, the hyperbolic tangent, scaled optimally according to \cite{lecun2012efficient}. Note, that unlike \cite{ororbia2017learning}, each specific pre-activation term no longer requires the original biases or scales due to the use of layer normalization. Since the Delta-RNN takes advantage of parameter-sharing, it requires substantially fewer layer normalizations than the GRU model, described next.

\subsection{The Multi-User Gated Recurrent Unit}
\label{gru_model}
The GRU variant of our synthesis model is defined as follows:
\begin{align}
\mathbf{z}_t &= \begin{aligned}[t]
\sigma(&LN(W_z \mathbf{x}_t) + LN(U_z \mathbf{e}_j)\\
&        + LN(V_z \mathbf{h}_{t-1}))
\end{aligned}\\
\mathbf{r}_t &= \sigma(LN(W_r \mathbf{x}_t) + LN(U_r \mathbf{e}_j) + LN(V_r \mathbf{h}_{t-1}))\\
\mathbf{\widehat{h}}_t &= \begin{aligned}[t]
\tanh(&LN(W_{\widehat{h}} \mathbf{x}_t) + LN(U_{\widehat{h}} \mathbf{e}_j)\\
&        + LN(V_{\widehat{h}} (\mathbf{r}_t \otimes \mathbf{h}_{t-1})))
\end{aligned} \\
\mathbf{h}_t &= \mathbf{z}_t \otimes \mathbf{h}_{t-1} + (1 - \mathbf{z}_t) \otimes \mathbf{\widehat{h}}_{t}
\end{align}
The architecture requires considerably more layer normalization operations than the Delta-RNN (and thus many more scale and shift vector biases). This is due to the fact that the GRU requires computing many variations of the linear pre-activations in order to make use of its various gates.